\documentclass[sigconf, nonacm]{acmart}

    \usepackage{kantlipsum}
    \usepackage[capitalise,noabbrev]{cleveref} 
    \usepackage{csquotes} 
    \usepackage{subcaption} 
    \usepackage{wrapfig} 
    \usepackage{booktabs} 
    \usepackage{multirow} 

    \usepackage{amsmath}    
    \usepackage{amsthm}     
    
    \DeclareMathOperator*{\argmin}{arg\,min}
    
    \usepackage{algorithm}  
    \usepackage{algorithmic} 

    \usepackage{fvextra}

    \DefineVerbatimEnvironment{PromptVerbatim}{Verbatim}{
      fontsize=\small, 
      baselinestretch=0.9,
      xleftmargin=1em,
      commandchars=\\\{\},
      breaklines=true,
    }

    \usepackage[edges]{forest}

    \usepackage{enumitem}

    \usepackage{multicol}

\begin{document}

\title{Learning from Anonymized and Incomplete Tabular Data}

\author{Lucas Lange}
\affiliation{%
  \institution{Leipzig University \& ScaDS.AI}
  \city{Leipzig}
  \country{Germany}
}
\email{lange@informatik.uni-leipzig.de}
\orcid{0000-0002-6745-0845}

\author{Adrian Böttinger}
\affiliation{%
  \institution{Leipzig University \& ScaDS.AI}
  \city{Leipzig}
  \country{Germany}
}
\email{a.boettinger@t-online.de}
\orcid{}

\author{Victor Christen}
\affiliation{%
  \institution{Leipzig University \& ScaDS.AI}
  \city{Leipzig}
  \country{Germany}
}
\email{christen@informatik.uni-leipzig.de}
\orcid{0000-0001-7175-7359}

\author{Anushka Vidanage}
\affiliation{%
  \institution{The Australian National University}
  \city{Canberra}
  \country{Australia}
}
\email{anushka.vidanage@anu.edu.au}
\orcid{0000-0002-5386-5871}

\author{Peter Christen}
\affiliation{%
  \institution{The Australian National University}
  \city{Canberra}
  \country{Australia}
}
\email{peter.christen@anu.edu.au}
\orcid{0000-0003-3435-2015}

\author{Erhard Rahm}
\affiliation{%
  \institution{Leipzig University \& ScaDS.AI}
  \city{Leipzig}
  \country{Germany}
}
\email{rahm@uni-leipzig.de}
\orcid{0000-0002-2665-1114}

\begin{abstract} 
User-driven privacy allows individuals to control whether and at what granularity their data is shared, leading to datasets that mix original, generalized, and missing values within the same records and attributes. While such representations are intuitive for privacy, they pose challenges for machine learning, which typically treats non-original values as new categories or as missing, thereby discarding generalization semantics. For learning from such tabular data, we propose novel data transformation strategies that account for heterogeneous anonymization and evaluate them alongside standard imputation and LLM-based approaches. We employ multiple datasets, privacy configurations, and deployment scenarios, demonstrating that our method reliably regains utility. Our results show that generalized values are preferable to pure suppression, that the best data preparation strategy depends on the scenario, and that consistent data representations are crucial for maintaining downstream utility. Overall, our findings highlight that effective learning is tied to the appropriate handling of anonymized values.
\end{abstract}

\maketitle

\vspace{.3cm}
\begingroup\noindent\raggedright\textbf{Availability:}
The source code and reference data for this paper are available at \url{https://github.com/luckyos-code/user-driven-privacy}.
\endgroup
\vspace{4.4cm}

\section{Introduction}\label{sec:intro}

Growing awareness of data privacy is changing how personal data is collected, shared, and analyzed~\cite{Ernst2016TheIO,song2021predictors,lange2025slice,gdpr2016}. Instead of passively ceding control, users increasingly demand transparency and agency over how their data is used. This shift has led to decentralized paradigms such as Personal Online Datastores (PODs)~\cite{solid2016,mansour2016demonstration} and federated learning~\cite{mcmahan2017communication}, where data privacy decisions are driven by individual users rather than a central authority.

As a practical example, we are involved in a collaborative project
with health researchers
to re-invent the process of conducting clinical trial research
using PODs. We are developing a participant-centered model for data
ownership with a trial being prepared to run (starting in February
2026) where sensitive medical and personal data is collected and
managed in each participant's own POD.
This novel method of conducting clinical research allows a trial to
be conducted remotely and at scale, supporting rapid testing,
validation, and accelerated collection of research results for
regulatory approval. By employing PODs, with participants being
involved in sharing their data with researchers, clinical trials
can now respect and preserve each participant's data privacy.

User-driven privacy can alter the structure of collected datasets, as rather than applying a single anonymization policy globally, users may selectively disclose information at different levels of granularity and sensitivity~\cite{solid2016,mansour2016demonstration}. As \cref{tab:example} shows, such datasets can contain heterogeneous mixtures of original, generalized (e.g., [30--39], old, Europe), and fully suppressed or missing values ($\bot$). While such representations provide intuitive privacy controls and can increase user trust~\cite{Ernst2016TheIO}, they reduce feature specificity and introduce ambiguity, posing challenges for downstream utility~\cite{mohammed2025dq}.

\setlength{\columnsep}{8pt}
\begin{wraptable}{r}{3.3cm}
\vspace{-\intextsep}
\setlength{\tabcolsep}{3.5pt}
\centering
\caption{Granular data.}
\label{tab:example}
\begin{tabular}{c|c|c}
Age & Sex & Country \\
\hline
26 & m & France \\
{[40--49]} & f & $\bot$\\
old & $\bot$ & Europe \\
\end{tabular}
\vspace{-\intextsep}
\end{wraptable}

Conventional Machine Learning (ML) pipelines alone are ill-equipped to handle this setting, and standard approaches to missing data treat absent values as stochastic noise or errors~\cite{rubin1976missing}, whereas generalized values encode correct but deliberately coarsened information with explicit semantic constraints. Treating generalized values as missing discards this information and can unnecessarily degrade performance. At the same time, privacy-preserving techniques for ML data such as $k$-anonymity~\cite{sweeney2002k}, differential privacy~\cite{dwork2006differential}, and synthetic data generation~\cite{lange2024generating} do not address learning from datasets with record-level mixed privacy transformations.
These issues become particularly pronounced in realistic deployment scenarios, where anonymization may affect data during training, inference, or both. Inconsistent data representations between training and inference can induce severe distribution shifts~\cite{mohammed2025dq,khan2025shades}.

In this work, we adopt a data-centric perspective and treat learning under user-driven privacy as a data preparation problem, focusing on refining such data representations.
We assess how different preparation strategies affect downstream ML performance and gain insights into handling anonymized tabular data for utility.

\noindent\textbf{Our Contributions:}
(1)~We formalize the problem of learning from anonymized tabular data, where original, generalized, and missing values co-exist.
(2)~We empirically evaluate data preparation methods across different scenarios to address this problem.
(3)~Our granularity-based imputation through specialization leverages generalization semantics and achieves strong utility and robustness.
(4)~We find that (i) generalization outperforms pure suppression, (ii) data preparation effectiveness is scenario-dependent, and (iii) consistent data preparation is critical for downstream utility.
\section{Related Work}
\label{sec:background}

In this section, we provide context for our problem and contributions by summarizing related work. Our work addresses challenges at the intersection of data quality, privacy, and machine learning.

\textbf{Personal Data and User-Driven Privacy:}
    Traditional privacy mechanisms typically apply uniform transformations across entire datasets, such as syntactic privacy models like $k$-anonymity~\cite{sweeney2002k} and its extensions~\cite{machanavajjhala2007diversity,li2007t}, which enforce indistinguishability by globally suppressing or generalizing attributes until group-based constraints are satisfied~\cite{lefevre2006mondrian}. Differential privacy~\cite{dwork2006differential} follows a different paradigm by injecting calibrated noise into queries or model updates to bound the influence of individual records. While both approaches provide formal privacy guarantees, their parameters are difficult for non-expert users to interpret, and privacy decisions are typically applied centrally rather than individually~\cite{Ernst2016TheIO}. Personalized versions of these mechanisms have been based on sensitivity or user preference but are either applied to the ML model or by excluding users from a dataset~\cite{boenisch2023have,lange2025sampling,jorgensen2015conservative,xu2020federated,schneider2024distributed}.
    
    Decentralized approaches such as PODs~\cite{solid2016,mansour2016demonstration} and federated learning~\cite{mcmahan2017communication} shift privacy control to users, allowing them to decide how precisely their data is shared. Empirical studies show that privacy preferences vary substantially across individuals and contexts~\cite{song2021predictors}, leading to datasets in which records differ in granularity across attributes. While such user-driven privacy can increase transparency and, in some cases, utility through less strict users~\cite{acquisti2016economics}, it also produces heterogeneous representations that are not addressed by existing privacy-aware ML pipelines. Prior work on PODs has largely focused on system architecture and access control~\cite{mortier2016personal}, leaving the implications for downstream ML largely unexplored. It is important to note that, unlike traditional privacy mechanisms, user-driven approaches typically lack formal privacy guarantees.

\textbf{Data Quality for Machine Learning:}
    Data-centric ML emphasizes that improving data quality often yields greater benefits than improving models~\cite{zha2025data,whang2023data}. In this context, studies~\cite{li2021cleanml,mohammed2025dq} confirm that missing values and feature inaccuracies have a strong impact on ML performance, and that these effects depend on whether data quality issues affect training data, inference data, or both. We share all three concerns, with deployment scenarios, feature completeness, and feature accuracy being key factors for our problem. However, our setting differs in an important respect: heterogeneity is intentional rather than erroneous. Generalized and missing values arise from user-driven privacy choices and encode partial but valid information. The challenge, therefore, is not to correct false data toward an assumed ground truth but to prepare anonymized data in a way that preserves utility while leveraging the inherent residual value of privacy constraints.
    
\textbf{Missing Value Imputation:}
    Missing data is commonly formalized using the framework by \citet{rubin1976missing}, which models missingness as a stochastic process. Classical imputation techniques replace missing values with point estimates, such as mean/mode substitution~\cite{jamshidian2007advances} and multivariate imputation~\cite{white2011multiple}. More recent approaches seen in benchmarks~\cite{miao2022experimental,khan2025shades} use learned models, including autoencoders~\cite{nazabal2020handling,gondara2018mida,mattei2019miwae} and generative models~\cite{li2019misgan,yoon2018gain,ipsen2021notmiwae} that infer plausible values from observed data. Across these methods, missing values are typically treated as noise or incomplete observations.
    
    Our setting differs fundamentally in that intentional abstraction is not the same as missing data. Generalized values encode interval- or set-valued constraints (e.g., age $\in [30$--$39]$) that preserve partial but semantically meaningful information. Treating such values as unconstrained missingness discards these constraints and can lead to implausible imputations. Related work on censored, interval-valued, or hierarchical data~\cite{klein2003survival,wang2015learning,speidel2020hmi} models coarse observations as uncertainty over a latent exact value, assuming a uniform observation mechanism, and targets valid statistical inference for a fixed analysis model. In contrast, we address mixed granularity within columns and records, focusing on preparation strategies that regain utility rather than posterior recovery.
    
    In a recent large-scale evaluation, \citet{khan2025shades} further highlight the limitations of imputation-centric approaches. They show that no imputation method consistently outperforms others across settings, that imputation effectiveness depends on whether missingness affects training data, inference data, or both, and that common imputation quality metrics are poor predictors of downstream ML performance. Our work builds on these insights by directly addressing the impact on ML utility and by studying preparation strategies under varying training and inference conditions.
    
    Finally, related work on data cleaning uses ML to detect and repair erroneous or incomplete records. Systems such as Lopster~\cite{reis2024lopster}, HoloClean~\cite{rekatsinas2017holoclean}, and CPClean~\cite{karlas2020cpclean} assume the existence of a ground truth toward which data should be corrected, often requiring clean training data or external constraints. In contrast, generalized values in our setting are not errors but deliberate abstractions that still encode partial semantic information, and no clean reference data is available. Rather than repairing values, our goal is to prepare anonymized data representations in a way that utilizes this information for downstream learning.

\textbf{Large Language and Foundation Models:}
    Large Language Models (LLMs) have been applied to imputation, error detection, and prediction on tabular data~\cite{narayan2022can,hegselmann2023tabllm,zhang2024llmpreprocessors,huang2024missing,wang2025llm,fang2024large}. More recently, Tabular Foundation Models (TFMs) are designed specifically for structured data~\cite{vanbreugel2024position}, with a debate on whether specialized TFMs or general-purpose LLMs are better suited for data management~\cite{papotti2025panel}. TabPFN~\cite{hollmann2025tabpfn} is a TFM that achieves strong predictive performance through in-context learning and without task-specific training. However, these models typically assume homogeneous feature representations and conventional missingness patterns. Thus, TabPFN can suffer under data quality issues, with errors distorting results for both corrupted and clean samples~\cite{papastergios2025qualitab}.

    In most cases, LLMs and TFMs treat generalized values as missing or categorical tokens, thereby discarding semantically meaningful information. Models may exploit such context when it is explicitly encoded and can rely on prior knowledge from their general pre-training. They, however, incur substantial computational costs.
    
    For a fair comparison across methods, we consider zero-shot LLM-based methods, i.e., without fine-tuning beyond the given prompt context, as we directly work on anonymized data only.
\section{Problem Formulation}\label{sec:problem}

We formulate the problem of learning from granular privatized data. We then present our solutions in \cref{sec:methodology} and describe the experimental environment for their evaluation in \cref{sec:experiments}.

\subsection{Data Model and Notation}

    Let $D = \{(x_i, y_i)\}_{i=1}^n$ denote a dataset with $n$ records, where $x_i \in \mathcal{X}$ is a feature vector and $y_i \in \mathcal{Y}$ the label for record $i$. Each feature vector $x_i = (x_i^{(1)}, \ldots, x_i^{(d)})$ consists of $d$ attributes, which may be numerical (e.g., age, income) or categorical (e.g., sex, occupation).
    
    \textbf{Privacy Transformations:}
    In user-driven privacy systems, each user $u_i$ controlling record $i$ applies a privacy transformation $\pi_{u_i}: \mathcal{X} \rightarrow \mathcal{X}^A$ to their record before its addition to a dataset. We consider three types of transformations per attribute:
    
    \begin{enumerate}[leftmargin=*]
        \item Original ($\pi^{\text{orig}}$): Remains unchanged, $\pi^{\text{orig}}(x_i^{(j)}) = x_i^{(j)}$
        
        \item Generalized ($\pi^{\text{gen}}$): Is replaced with a less specific value:
        \begin{itemize}[leftmargin=0pt]
            \item \emph{Numerical}: Replaced by range, e.g., $\pi^{\text{gen}}(35) = [30$--$39]$
            \item \emph{Categorical}: Replaced by group, e.g., $\pi^{\text{gen}}(\text{``PhD''}) = \text{``graduate''}$
        \end{itemize}
        
        \item Missing ($\pi^{\text{miss}}$): Value is removed, $\pi^{\text{miss}}(x_i^{(j)}) = \bot$
    \end{enumerate}
    
    Users may apply any transformation to any attribute. Let $\tau_i \in \{\text{orig}, \text{gen}, \text{miss}\}^d$ denote the transformation type vector for record $i$, where $\tau_i^{(j)}$ indicates the transformation applied to attribute $j$ of record $i$, resulting in the anonymized dataset $D^A = \{(\pi_{u_i}(x_i), y_i)\}_{i=1}^n$.
    
    \textbf{Privacy Configurations:}
    We characterize datasets by the distribution of transformations across records. A privacy configuration tuple $(p_{\text{orig}}, p_{\text{gen}}, p_{\text{miss}})$ specifies the fraction of attribute-record pairs in each transformation state, where $p_{\text{orig}} + p_{\text{gen}} + p_{\text{miss}} = 1$. 
    
    For readability, we denote configurations using the shorthand \texttt{orig-gen-miss} notation.
    For example, \texttt{66-17-17} indicates 66\% original values, 17\% generalized, and 17\% missing.

\subsection{Learning Objective and Data Preparation}
    Given anonymized dataset $D^A = D^A_{\text{train}} \cup D^A_{\text{test}}$, the goal is to learn a model $f: \mathcal{X}^A \rightarrow \mathcal{Y}$ that minimizes the expected prediction error: $\mathcal{L}(f) = \mathbb{E}_{(x,y) \sim D^A_{\text{test}}}[\ell(f(x), y)]$, where $\ell$ is a loss function (e.g., 0-1 loss for classification). The model $f$ is trained exclusively on the anonymized $D^A_{\text{train}}$ without access to original data other than what can be observed directly from values in $D^A$ itself.
    
    \textbf{Challenges for Standard ML:}
    Traditional ML assumes $D_{\text{train}}$ and $D_{\text{test}}$ are drawn from the same distribution with consistent feature representations. Our data model with heterogeneous anonymization violates these assumptions, posing several challenges. We focus on three key aspects: (1)~\emph{feature ambiguity}, where generalized values such as age interval [30--39] represent multiple possible true values, creating uncertainty in the feature space, (2)~\emph{information loss}, since generalized and missing values reduce the mutual information between features and labels compared to original data, and (3)~\emph{semantic dilution}, where the feature domain is widened by generalized values that semantically still contain information about the original value, which standard ML algorithms cannot harness.
    
    \textbf{Data Preparation:}
    To address these challenges, we introduce a data preparation function $\phi: \mathcal{X}^A \rightarrow \mathcal{X}^P$ that maps anonymized features of a record into a representation potentially more suitable for ML. Such a function $\phi$ can be applied to training data, test data, or both, yielding: $D^P = \{(\phi(x_i), y_i) \mid (x_i, y_i) \in D^A\}$.
    The data preparation goal is to find a $\phi$ that minimizes the utility gap between using prepared anonymized data and the original data:
    $\phi^* = \argmin_{\phi \in \Phi} \left| \mathcal{U}(f^*) - \mathcal{U}(f_\phi) \right|$, where:
    
    \begin{itemize}[leftmargin=*, topsep=3pt]
        \item $f^*$ is a baseline classifier trained and evaluated on original data,
        \item $f_\phi$ is a classifier trained on $\phi(D^P_{\text{train}})$ and evaluated on $\phi(D^P_{\text{test}})$,
        \item $\mathcal{U}(\cdot)$ measures model utility (e.g., accuracy, F1-score, etc.),
        \item $\Phi$ is the space of possible preparation methods.
    \end{itemize}
    
    An ideal preparation method achieves $\mathcal{U}(f_\phi) \approx \mathcal{U}(f^*)$, effectively maintaining predictive performance despite anonymization.
    
    \subsection{Deployment Scenarios:}
    Preparation requirements change as anonymization may affect data for training, inference, or both. Data transformations should generally be applied consistently to avoid distribution shift, as inconsistent preparation ($\phi_{\text{train}} \neq \phi_{\text{test}}$) results in mismatched feature distributions. However, when only one set is anonymized, preparation may cause information loss for the other. We identify three deployment scenarios that represent distinct real-world contexts:
    
    \begin{enumerate}[leftmargin=*,itemsep=2pt]
        \item \emph{Anonymized training (AnTr)}: Training data is sensitive and anonymized, but inference occurs in a trusted environment with access to original values (e.g., training on anonymized patient records, deploying in a secure hospital system).
        
        \item \emph{Anonymized inference (AnTe)}: Training uses public or consented data, but inference handles privacy-conscious users who routinely anonymize their inputs (e.g., a model trained on census data, queried by users who generalize sensitive attributes).
        
        \item \emph{Both anonymized (AnBo)}: Privacy constraints apply throughout, requiring consistent handling of anonymized data in both phases (e.g., federated learning with user-controlled privacy).
    \end{enumerate}
    
    In (1) and (2), data preparation is applied only to the anonymized part, whereas in (3), both require consistent preparation methods.

\begin{figure*}[t]
\centering
\begin{forest}
  forked edges,
  for tree={
    grow=south,
    draw,
    rounded corners,
    align=center,
    edge={->,>=stealth},
    l sep=0.6cm,
    s sep=0.3cm,
    anchor=north,
    fork sep=0.3cm,
  },
  where level=2{l sep=0.4cm}{}
  [Anonymized Dataset
    [No Preparation\\{\small Anonymized Baseline}]
    [Standard Imputation\\{\small Missing Values}
      [Simple]
      [(C-)MICE]
    ]
    [Granularity-Based Imputation\\{\small Generalized Values}
      [Generalization]
      [Specialization]
    ]
    [LLM-Based\\{\small Zero-Shot}
      [Imputation]
      [Prediction]
    ]
  ]
\end{forest}
\caption{Overview of evaluated preparation methods for anonymized datasets with generalized and missing values.}
\label{fig:method_overview}
\end{figure*}
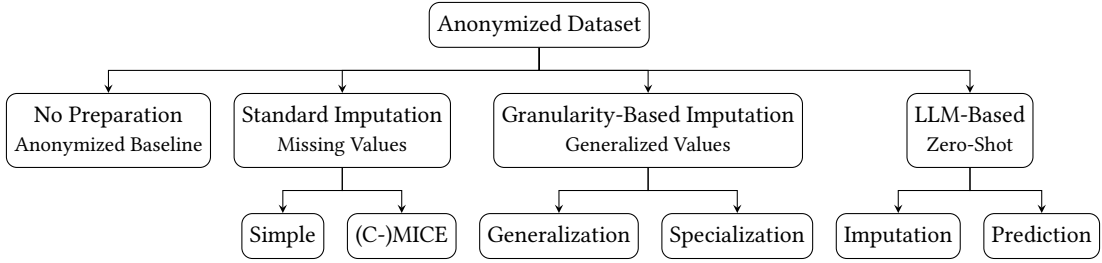

\begin{figure*}[t]
\centering
\begin{subfigure}[t]{0.14\textwidth}
\raggedleft
\vspace{3.3em}
Original value: \\
Generalized value: \\
Missing value: \\
\end{subfigure}
\hfill
\begin{subfigure}[t]{0.21\textwidth}
\centering
\caption{Anonymized Dataset}
\begin{tabular}{c|c|c|c}
ID & Age & Sex & $y$ \\
\hline
R1 & 52 & w & 1 \\
R2 & {[}50--59{]} & m & 0 \\
R3 & 42 & $\bot$ & 1 \\
\end{tabular}
\end{subfigure}
\hfill
\begin{subfigure}[t]{0.21\textwidth}
\centering
\caption{Standard Imputation}
\begin{tabular}{c|c|c|c}
ID & Age & Sex & $y$ \\
\hline
R1 & 52 & w & 1 \\
R2 & 42 & m & 0 \\
R3 & 42 & m & 1 \\
\end{tabular}
\end{subfigure}
\hfill
\begin{subfigure}[t]{0.21\textwidth}
\centering
\caption{Generalization}
\begin{tabular}{c|c|c|c}
ID & Age & Sex & $y$ \\
\hline
R1 & {[}50--59{]} & w & 1 \\
R2 & {[}50--59{]} & m & 0 \\
R3 & {[}40--49{]} & $\bot$ & 1 \\
\end{tabular}
\end{subfigure}
\hfill
\begin{subfigure}[t]{0.21\textwidth}
\centering
\caption{Specialization ($\tilde{v} \geq 2$)}
\begin{tabular}{c|c|c|c||c}
ID & Age & Sex & $y$ & $w$ \\
\hline
R1 & 52 & w & 1 & \textit{1.0} \\
R2 & 55 & m & 0 & \textit{1.0} \\
R3 & 42 & w & 1 & \textit{0.5} \\
R3 & 42 & m & 1 & \textit{0.5} \\
\end{tabular}
\end{subfigure}
\caption{Comparison of possible preparation outcomes on an anonymized data sample, where $\bot$ denotes missing values.}
\label{fig:method_comparison}
\end{figure*}

\section{Data Preparation Methods}
\label{sec:methodology}

We now propose a set of data preparation methods designed to handle heterogeneously anonymized tabular datasets. As illustrated in \cref{fig:method_overview}, we group these techniques into categories according to their core operating principles. Each category reflects a specific strategy for navigating the trade-off between data granularity and utility. Their effects on the anonymized data are shown in \cref{fig:method_comparison}.

\subsection{No Preparation (Anonymized Baseline)} 

This baseline feeds anonymized data directly to the ML model without alterations. The drawback of this solution is that varying granularities in column domains make it harder for a model to correctly align representations with labels during training. This is particularly problematic when only either the training or test data is anonymized since there will be a mismatch between learned and predicted statistics. When both sets use similar anonymization patterns, the model might be able to learn appropriate mappings.

\subsection{Standard Imputation Methods}

As a common solution and comparison point to our other methods, we use standard statistical imputation strategies~\cite{little2019statistical}. These statistical techniques learn to impute plausible replacements from patterns in the available data but they focus on missing values in datasets. Directly applying them to our data thus treats generalized transformations as missing data.

\textbf{Simple Imputation:}
Replaces missing values with the attribute's mode (categorical) or mean (numeric) calculated from the remaining original values~\cite{jamshidian2007advances}. This approach is fast and simple but ignores inter-attribute correlations and can distort feature distributions.

\textbf{MICE:}
(Multivariate Imputation by Chained Equations)~\cite{white2011multiple} models each attribute as a function of all others using iterative regression. For each attribute with missing values it (1) trains a predictive model using records with observed values, (2) imputes missing values conditioned on other attributes, and (3) iterates until convergence. MICE thereby captures inter-attribute dependencies and produces statistically plausible values. However, it can violate generalization constraints since it has to assume a missing value for generalized entries, which leads, for example, to imputing age $=25$ for a record where the user specified age $\in[30$--$39]$, thereby ignoring the semantic constraints encoded in the generalized value. 

\textbf{Constrained MICE (C-MICE):}
To incorporate the available information from generalized values, we augment MICE with post-processing that clips imputed values to their generalization domains. For example, if MICE imputes age $=25$ for a record with age $\in[30$--$39]$\footnote{During preparation, interval categories like ``old'' are handled as a range in all methods.}, we replace it with $30$. This exploits semantic bounds but can create distribution artifacts at generalization set boundaries.

\subsection{Granularity-Based Imputation Methods}

Unlike baseline imputation, our granularity-based imputation works directly with the hierarchical structure of generalized values. These methods aim to unify the granularity levels inside the columns and between training and test data by either generalizing all data or expanding generalized values to their possible originals.

\textbf{Forced Generalization:}
This method creates consistency within the column domains by raising all original values or values of lower granularity to a common granularity level, while missing values remain unchanged. This approach, formalized in \cref{alg:generalization}, efficiently creates a homogeneous representation but discards more detailed information from original and less generalized values.

\textbf{Weighted Specialization with Filtering:}
Opposed to the forced generalization, specialization inverts generalizations by expanding each generalized value to its possible originals, as observed in the anonymized column data. Since missing values are nothing else than "extremely generalized" values, we also apply specialization to them. This is similar to common imputation methods but we aim to incorporate the granular nature of a dataset into the method design. For this,  we have to address two key challenges that occur with this data preparation approach: (1) exponential dataset growth from the Cartesian product of possible values for each generalized or missing entry and (2) equal treatment of candidates with varying plausibility. The algorithm is given in \cref{alg:specialization} and we explain its workings through its parts: specialization, weighting, and filtering.

\textit{Specialization:} For each anonymized record, we generate all concrete value combinations consistent with its generalized attributes. Possible values are restricted to those observed in the corresponding anonymized dataset column. Numeric attributes are instantiated using the range midpoint (e.g., age $\in[30$--$39] \rightarrow 35$), while categorical attributes are expanded to all values within the generalization (e.g., ``Higher Education'' $\rightarrow$ \{``Bachelors'', ``Masters'', ``Doctorate''\}, or for the sex column with missing value ``$\bot$'' $\rightarrow$ \{``m'', ``f''\}). 
While this approach can be applied directly, it has two critical shortcomings: (1) if $c$ categorical attributes are generalized into groups of average size $g$, a single record expands into $g^c$ specialized variants, leading to a combinatorial dataset growth of up to $O(n \cdot g^c)$; and (2) because generalized records are expanded into many variants while unspecialized records remain singletons, the resulting dataset exhibits a strong distribution skew in favor of specialized entries.

\textit{Filtering:} To limit the scope of the first problem and control dataset size, we only keep the $\tilde{v}$ most relevant variants per record immediately after expanding into the possible variants. This bounds the output size, which is reduced from $O(n \cdot g^c)$ to $O(n \cdot \tilde{v})$. For filtering the most relevant variants, we use a ranking strategy: We score variants by similarity to column profiles $\mathcal{P}$ built from observable original dataset values and select the top-$\tilde{v}$ variants. We score each variant $s'$ as: $\text{score}(s') = \frac{1}{d}\sum_{j=1}^{d} \text{sim}_j(s[j], \mathcal{P}_j)$, where $d$ is the number of attributes and $\text{sim}_j$ returns the normalized frequency for categorical attributes or the z-score similarity as $\max(0, 1 - |z_j|/3)$ for numeric attributes, with $z_j$ being the z-score relative to the profile mean and standard deviation. This leads to higher scores for variants with common value combinations in the original values.

\begin{algorithm}[t]
\caption{Forced Generalization}\label{alg:generalization}
\begin{small}
\begin{algorithmic}[1]
\REQUIRE Dataset $D^A$, generalization functions $\{g^j_\ell\}$ for each attribute $j$ and level $\ell$
\ENSURE Generalized dataset $D^P$
\STATE $D^P \gets D^A$
\FOR{each attribute $j \in \{1,\ldots,d\}$}
    \STATE $\ell \gets$ highest level in hierarchy for attribute $j$
    \FOR{each record feature vector $x_i \in D^P$}
        \IF{$x_i[j]$ is not missing}
            \STATE $x_i[j] \gets g^j_\ell(x_i[j])$ \COMMENT{Generalize to level $\ell$}
        \ENDIF
    \ENDFOR
\ENDFOR
\RETURN $D^P$
\end{algorithmic}
\end{small}
\end{algorithm}

\textit{Weighting:} The second problem, distribution skew, is addressed by introducing record-level weights. Original records retain a unit weight of $w=1$, while each derived variant is assigned a weight of $w = 1/|S_i|$, where $|S_i| \leq \tilde{v}$ denotes the number of variants retained after filtering for record $i$. This normalization ensures that each record from the anonymized input dataset contributes the same total weight, whether it appears once (i.e., is not specialized) or is expanded into multiple specialized variants. 

During ML training, these weights are incorporated via sample weighting, reducing the influence of individual variants when $w<1$. As a result, specialization-induced record inflation does not disproportionately affect the learned model.
With specialization on test data, we may generate multiple candidate records from a single anonymized record. During inference, this leads to multiple predictions for an individual, which then need to be aggregated into a final decision for the original entry. This is done by calculating the weighted mean of all predictions from specialized entries: $p_{\text{avg}} = \sum_{i=1}^{\tilde{v}} w_i p_i \,/\, \sum_{i=1}^{\tilde{v}} w_i$, where $\{p_1, \ldots, p_{\tilde{v}}\}$ are the predictions from $\tilde{v}$ specialized entries with weights $\{w_1, \ldots, w_{\tilde{v}}\}$. In this way, we obtain a standard inference output independent of variant-split record inputs. The weights fulfill the same function as in training, leading to predictions that equally consider all candidates.

\textit{Without Original Values:}
For fully anonymized datasets without any original values left, we can neither construct profiles nor have any observed original values to sample from for categorical features. Thus, we cannot create categorical variants and must keep their generalized versions. For numerical features, we can still use the midpoints of generalized interval values. For missing numerical values, we can utilize the boundaries of all generalized intervals for this feature to assume the overall bounds and resulting midpoints.

\begin{algorithm}[t]
\caption{Weighted Specialization with Filtering}\label{alg:specialization}
\begin{small}
\begin{algorithmic}[1]
\REQUIRE Dataset $D^A$, generalization mappings $\{G_j\}$, number of\\variants kept after filtering $\tilde{v}$
\ENSURE Specialized dataset $D^P$ with weights $W$
\STATE $D^P \gets \emptyset$, $W \gets \emptyset$
\STATE Build column profiles $\mathcal{P}$ from original values in $D^A$ if any
\FOR{each record feature vector $x_i \in D^A$}
    \STATE $S_i \gets \{x_i\}$ \COMMENT{Candidates for this record}
    \FOR{each attribute $j \in \{1,\ldots,d\}$}
        \STATE $S'_i \gets \emptyset$
        \FOR{each entry $s \in S_i$}
            \IF{$s[j]$ is original}
                \STATE Add $s$ to $S'_i$
            \ELSIF{$s[j]$ is numeric and generalized}
                \STATE $s[j] \gets \text{midpoint of generalized range in s[j]}$
                \STATE Add $s$ to $S'_i$
            \ELSIF{$s[j]$ is numeric and missing}
                \STATE $s[j] \gets \text{midpoint of total $s[j]$ range in $D^A$}$
                \STATE Add $s$ to $S'_i$
            \ELSIF{$s[j]$ is categorical and generalized}
                \STATE $V \gets G_j(s[j])$ \COMMENT{Based on original values in $D^A$}
                \FOR{each $v \in V$}
                    \STATE $s' \gets \text{copy}(s)$; $s'[j] \gets v$
                    \STATE Add $s'$ to $S'_i$
                \ENDFOR
            \ELSIF{$s[j]$ is categorical and missing}
                \STATE $V \gets$ \text{all possible original values for $s[j]$ in $D^A$}
                \FOR{each $v \in V$}
                    \STATE $s' \gets \text{copy}(s)$; $s'[j] \gets v$
                    \STATE Add $s'$ to $S'_i$
                \ENDFOR
            \ENDIF
        \ENDFOR
        \STATE $S_i \gets S'_i$
    \ENDFOR
    \STATE $S_i \gets \textsc{Filter}(S_i, \tilde{v}, \mathcal{P})$ \COMMENT{Filter to $\tilde{v}$ variants} 
    \STATE $w \gets 1 / |S_i|$ \COMMENT{Equal weight per kept variant}
    \FOR{each $s \in S_i$}
        \STATE Add $s$ to $D^P$; Add $w$ to $W$
    \ENDFOR
\ENDFOR
\RETURN $D^P, W$
\end{algorithmic}
\end{small}
\end{algorithm}

\subsection{LLM Methods}

LLMs trained on vast text corpora have internalized statistical patterns about real-world entities and attributes~\cite{narayan2022can}. We leverage this knowledge for imputation and prediction on our anonymized datasets, however, due to their heavy reliance on prior knowledge, these methods may struggle on new tasks or specialized domains.

\textbf{LLM Imputation:}
We prompt a pretrained general LLM with reasoning capabilities to impute anonymized values given some context on the anonymization. The prompt configuration is:

\newcommand{\MissingVal}{$\bot$}
\begin{small}
\begin{PromptVerbatim}
\textcolor{blue}{'system_prompt':} "You are a data analyst filling missing or generalized values.",
\textcolor{blue}{'prompt_template':} """Dataset: \textcolor{blue}{\{dataset_name\}}
Task: Impute specific concrete values for the indicated columns in each record.

Context:
- "\MissingVal" means value is completely missing
- Values like [30–39] are generalized ranges
- Semantic values like "young" are generalizations

Instructions:
- For each record, provide a specific concrete value for EVERY column listed in "Targets".
- Predict specific values (e.g., "35" instead of "[30–39]", "Private" instead of "private_sector").

Records to process: \textcolor{blue}{\{records_block\}}

Instructions for output format:
- Return ONE LINE PER RECORD in this exact format (no JSON, no markdown): REQ_ID<TAB>col1=value1|col2=value2|...
- Use the pipe character `|` to separate column predictions.
- Values may contain spaces but MUST NOT contain the `|` or tab characters.
- If a value is unknown, return the string "UNK" for that column.

Example lines:
REQ_0   age=35|workclass=Private
REQ_1   occupation=Sales

Return ONLY the lines, nothing else."""
\end{PromptVerbatim}
\end{small}

LLMs are a great fit for this purpose since they capture complex, non-linear patterns even in a zero-shot setting with no additional training required. These models can be fine-tuned but for sensitive data without ground-truth this would not be realistic.
However, this imputation method is limited by the expensive computational overhead from API access for a large amount of imputations in a dataset. This can be significant regarding monetary and time constraints, as well as environmentally problematic power usage.

\textbf{LLM Prediction:}
Instead of employing LLMs during the imputation step, we can also directly ask an LLM for predictions. In this approach, the LLM receives the test set records, which, depending on the case, contain generalized and missing values, and directly predicts the target label without any intermediate imputation efforts except for what is reasoned internally. In its prompt, the LLM receives anonymized records and tries to infer their target labels:

\begin{PromptVerbatim}
\textcolor{blue}{'system_prompt':} "You are predicting target variables.",
\textcolor{blue}{'prompt_template':} """Dataset: \textcolor{blue}{\{dataset_name\}}
Task: Predict the target variable '\textcolor{blue}{\{target_name\}}' (0 or 1) for the following records. \textcolor{blue}{\{target_info\}}

Instructions:
- Return ONE LINE PER RECORD in this exact format (no JSON, no markdown): REQ_ID<TAB>value
- value must be 0 or 1

Records to process: \textcolor{blue}{\{records_block\}}

Example:
REQ_1   0
REQ_2   1

Return ONLY the lines, nothing else."""
\end{PromptVerbatim}

The \texttt{target\_info} variable gives a brief input on the target variables, e.g. for Income: \enquote{Target values: 0 (<=50K income) or 1 (>50K income)}. As before, the LLM does not use pre-training on sensitive data and is instead used in a zero-shot setting. Since we directly predict target values, which is a less expensive task than imputing many column values on larger training sets, this method has less overhead compared to LLM imputation. However, contrasted against training a classic ML model for inference, we require many possibly expensive API requests and probably longer inference times. Another drawback of direct prediction is that there is even less information for the LLM regarding dataset-specific patterns especially in specialized domains, which can reduce its performance.

\section{Experimental Setup}
\label{sec:experiments}

\begin{table}[t]
    \centering
    \caption{Datasets; Cf. shows usage in current related work.}
    \label{tab:datasets}
    \small
    \begin{tabular}{@{}l l r r r r l@{}}
        \toprule
        \textbf{Name} & \textbf{Domain} & \textbf{\#Rows} & \textbf{\#Attrs} & \textbf{\#Cat} & \textbf{\#Num} & \textbf{Cf.} \\
        \midrule
        Credit & Finance & 1K & 20 & 13 & 7 & \cite{mohammed2025dq,khan2025shades} \\
        Income & Finance & 45K & 13 & 8 & 5 & \cite{reis2024lopster} \\
        Diabetes & Health & 70K & 21 & 14 & 7 & (\cite{khan2025shades}) \\
        Employment & Hiring & 381K & 16 & 15 & 1 & \cite{khan2025shades} \\
        \bottomrule
    \end{tabular}
\end{table}

This section describes the details of our experimental design.

\textbf{Datasets:}
We evaluate the public datasets in \cref{tab:datasets} from common domains in the context of privacy and fairness. The datasets vary in size and exhibit different ratios of categorical to numerical attributes. The German \textit{Credit} dataset~\cite{statlog_german_credit_data_144} is a smaller financial dataset with mixed features, focusing on creditworthiness prediction. The Adult Census \textit{Income} dataset~\cite{kohavi1996scaling} contains U.S. Census records with demographic attributes and predicts whether an individual earns more than \$50K annually. We further use the equal split binary version of the \textit{Diabetes} Readmission dataset~\cite{strack2014impact}, which is based on hospital records and predominantly consists of (binary) categorical medical features, aiming to infer a patient’s diabetes status. Finally, the Folk \textit{Employment} dataset~\cite{ding2021retiring}, also derived from U.S. Census data, comprises a large collection of mostly categorical applicant descriptors and predicts employment status. For all datasets, we focus on binary classification and use an 80/20 train-test split.

\textbf{Privacy Distributions:}
To simulate varying data distributions for our problem, as introduced in \cref{sec:problem}, we employ multiple configurations representing different user privacy and data quality preferences. In addition to the original data baseline, we examine cases where one third, two thirds, or all of the original columns is instead generalized or missing. To better study the influence of generalized data, we also include setups with only missing data. These distributions are specified as \texttt{Original-Generalized-Missing} percentages (\texttt{O-G-M}), resulting in the following list: \texttt{1-0-0} (baseline), \texttt{66-17-17}, \texttt{66-0-34}, \texttt{33-33-34}, \texttt{33-0-67}, and \texttt{0-66-34}. We include the setup without original data to examine which methods remain effective in this worst-case scenario. Instead of fully suppressing all values, we assign the remaining share to generalized values in order to retain some semantic information. However, this setup is in part evaluated separately since such an extreme case would heavily skew general results towards pre-trained methods like LLMs. 

\textbf{Privacy Simulation:}
To evaluate privacy distributions, we need to create anonymized dataset simulations through fitting transformations.
For that, we first have to define granularity levels for each attribute of our evaluation datasets.
The extremes on the granularity scale are original and missing, which either keep a value as it is or completely remove it.
In between these, there are generalization levels that semantically group a set of original values together.
These levels and their count can differ depending on what is possible with the feature; e.g., for age, we may define two levels, with the first being a range [50--59] for the original value 52 and the second a larger range like ``middle-aged''.
Transformations using these levels are then implemented by column-wise anonymization for each attribute based on the selected distribution.

Depending on the goal privacy distribution, e.g., \texttt{33-33-34}, for all attributes in a dataset, we randomly select 33\% of the values of an attribute to be kept original, 33\% to be generalized, and 34\% to be made missing.
If there are multiple possible generalization levels, one is selected with equal probability.
Applying these transformations results in a heterogeneous mixture of original, generalized, and missing values that mirrors the goal privacy distribution.
However, since binary attributes (e.g., sex) cannot be meaningfully generalized, they are instead treated as missing even when they should be generalized. This may cause deviations from the target configuration depending on the available features.
Finally, based on the scenarios given in \cref{sec:problem}, these transformations may be applied to either the training data, test data, or to both. 

\textbf{Performance and Privacy Metrics:}
The performance of each approach is evaluated using the macro F1-score~\cite{christen2023review}, which provides a balanced measure through the harmonic mean of per-class precision ($\text{P} = \frac{TP}{TP + FP}$) and recall ($\text{R} = \frac{TP}{TP + FN}$) averaged across classes C: $\text{F1} = \frac{1}{C} \sum_{c=1}^{C} 2 \cdot \frac{\text{P}_c \cdot \text{R}_c}{\text{P}_c + \text{R}_c}$. This metric was chosen over accuracy because it is not skewed by dataset class imbalance and treats all classes equally. However, we also consider macro precision and recall individually~\cite{christen2023review}.
Our key comparison is the performance retention relative to the original baseline, since our preparation should minimize the utility loss from anonymization. 
Accordingly, we do not rely on error-based metrics, as our methods do not specifically aim for exact reconstruction of the original data and, in some cases, do not even produce comparable representations.

Other than performance, we also need to assess the privacy attributes to see whether our preparation methods inadvertently reduce the privacy provided by the anonymization. We further need to test whether these user-driven transformations even provide additional privacy as they deliver no formal privacy guarantees. For that, we adopt established privacy metrics, with the first one being the \textit{marketer risk} ($R_m$)~\cite{dankar2010method} defined as the average probability that an attacker can uniquely re-identify a record based on quasi-identifiers: $R_m(D) = \frac{1}{|D|} \sum_{x_i \in D} \frac{1}{|\text{EC}(x_i)|}$, where $\text{EC}(x_i)$ is the equivalence class of record $x_i$ under quasi-identifier attributes $Q$, i.e., the set of records with identical values in all these features. Equivalently, $R_m = |\text{EC}| / |D|$ where $|\text{EC}|$ is the number of distinct ECs and lower $R_m$ is better.
As another baseline privacy metric, we use \textit{$k$-anonymity}~\cite{sweeney2002k}, which a dataset satisfies if every record is indistinguishable from at least $k{-}1$ other records: $\forall x_i \in D: |\text{EC}(x_i)| \geq k$. Here, we report the minimum EC size as an indicator of the maximum $k$ achieved for all records, where larger $k$ is better. For both metrics, we conservatively assume that $Q$ contains all attributes as sensitivity is user-defined and not predetermined.

\textbf{Implementation:}
We implemented our experiments in Python, using the XGBoost~\cite{chen2016xgboost} model as our ML pipeline, which is effective for tabular data and offers robustness to missing values with its tree structure~\cite{shwartz2021tabular}. However, LLM predictions are evaluated directly on test sets without the need for model training.
For reproducibility, we set all random seeds to 42 for the entire process, from deterministic data preparation to ML evaluation. Reproducibility also includes setting the LLM temperature to 0 for the imputation and prediction methods using the open-source reasoning model Mistral-Small-24B-Instruct-2501~\cite{mistralsmall3}.
All code is available on \url{https://github.com/luckyos-code/user-driven-privacy}, which also allows re-creating our anonymized datasets for further research purposes.

\begin{figure*}[t]
    \centering
    \includegraphics[width=\textwidth]{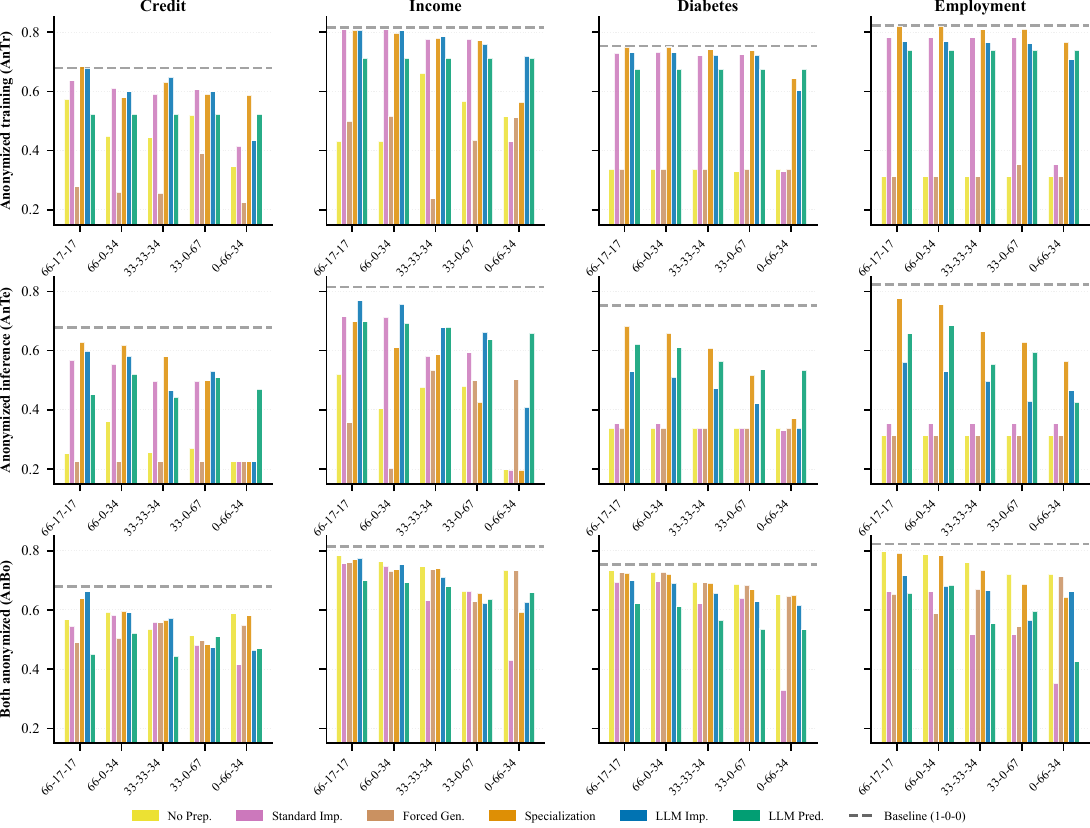}
    \caption{
        F1 scores for preparing methods across evaluation scenarios, datasets, and privacy distributions.
        The x-axis shows the distributions as Original-Generalized-Missing percentages.
        Dashed lines indicate baseline performance on original data. 
    }
    \label{fig:overview_combined}
\end{figure*}

\section{Results}
\label{sec:results}

We evaluate our preparation methods across four datasets spanning financial, medical, and employment domains, with baseline F1 scores of 0.68 (Credit), 0.81 (Income), 0.75 (Diabetes), and 0.82 (Employment) on the original data. \cref{fig:overview_combined} presents the complete results across all scenarios, datasets, and privacy configurations.

\subsection{Submethod Evaluation} 
    
    Both standard imputation and specialization comprise multiple submethods described in \cref{sec:methodology}.
    Therefore, we first identify their best-performing configurations for the subsequent evaluations.
    We compare methods using aggregated F1 performance, excluding \texttt{0-66-34}, as it contains no original values for imputation and would otherwise skew the results.
    For standard imputation, we evaluate three approaches: simple imputation, MICE, and C-MICE.
    Simple imputation achieves the highest F1 score (0.61), outperforming both MICE (0.59) and C-MICE (0.59), largely due to its stronger performance on the Income dataset (+0.13 F1).
    It is also more consistent across setups, being best in 35\% more cases than the MICE-based methods.
    This suggests that MICE may overfit to dataset-specific patterns, creating artificial label correlations that harm model training, whereas mode and mean replacement just reinforce existing feature statistics.
    For specialization, we evaluate the number of retained variant candidates per record ($\tilde{v} \in \{1, 2, 3, 5\}$).
    Profile-guided filtering with $\tilde{v}=2$ yields the best trade-off, achieving an aggregated F1 score of 0.69.
    It outperforms selecting one (0.68) or three (0.68) candidates, while raising to $\tilde{v}=5$ provides only marginal gains (+0.003 F1) but significantly increases dataset sizes.
    Based on these results, we select simple imputation and profile-guided specialization with $\tilde{v}=2$ for the remainder of our experiments.

\subsection{Anonymized Data and Privacy Distributions}

    We also want to examine the anonymized datasets created for evaluation.
    We generate these datasets using different privacy distributions, as described in \cref{sec:experiments}.
    Importantly, privacy transformations are applied column-wise according to the specified percentages.
    Thus, the share of original values refers to individual feature values rather than fully original records.
    Consequently, each record has the same probability of containing original, generalized, or suppressed features.
    This setup produces a more diverse and challenging dataset than approaches that enforce fixed ratios of fully original and fully anonymized records.
    By avoiding assumptions about user-level distributions, we obtain granular records throughout the dataset.
    In contrast, record-wise approaches may favor discarding anonymized records and retaining only fully original ones, thereby simplifying the problem to a narrower set of cases.
    
    As shown in \cref{fig:overview_combined}, we observe consistent utility degradation as anonymization increases, i.e., as the proportion of original values decreases.
    Moving from lighter (66\% original) to heavier anonymization (33\%), F1 scores decline gradually across all methods, while their relative ordering remains largely stable.
    A critical threshold is reached at \texttt{0-66-34}, where no original values remain. In this setting, even the strongest methods exhibit up to a 31\% relative decline compared to the original baseline.
    Across the remaining distributions, we find that despite the absence of fully original records, performance remains robust as long as a sufficient fraction of original values is present.
    Due to our column-wise anonymization, the share of fully original records is already low at 66\% original values (approximately 2.5\%) and drops to zero at 33\% and below.
    Finally, the presence of coarsened values in every record leads to variant generation for all entries under specialization, making dataset size an important consideration and motivating our filtering strategy.

\begin{figure}[t]
    \centering
    \includegraphics[width=0.9\linewidth]{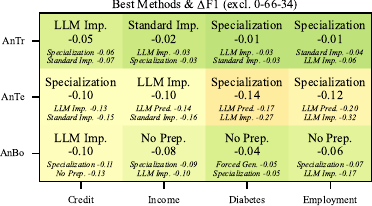}
    \caption{%
        Heatmap of average F1 utility loss against original baseline for the three best preparation methods across privacy distributions excluding \texttt{0-66-34} as a special case.
    }
    \label{fig:heatmap}
\end{figure}

\subsection{Scenario-Dependent Method Selection}

    Taking a closer look at the individual scenarios, \cref{fig:overview_combined} already indicates differences in how well methods perform across settings.
    The heatmap in \cref{fig:heatmap} reports the top three methods based on F1 retention relative to the original baseline, aggregated across datasets and privacy distributions while excluding \texttt{0-66-34}.
    We observe that no single preparation method dominates across all scenarios and datasets, although specialization consistently ranks among the top-performing approaches.
    Overall, the optimal method choice strongly depends on the deployment context.
    
    In Scenario AnTr, where models are trained on anonymized data but deployed on original data, performance remains close to the original baseline for most datasets, with Credit being the main exception.
    Scenario AnTe exhibits a clear performance decline across all datasets, even under the best-performing methods, making it the most challenging setting.
    This difficulty arises from the distribution shift between training on predominantly original values and inferring on coarsened inputs.
    Notably, direct LLM-based prediction, which bypasses task-specific model training, emerges as a competitive alternative in this scenario, ranking second for all datasets except Credit.
    This suggests that general world knowledge from pre-training can partially compensate for distribution mismatch.
    
    While Scenarios AnTr and AnTe benefit from data preparation, primarily through specialization, AnBo performs best when operating directly on anonymized data without any preparation.
    Here, consistent anonymization across training and inference removes distribution shift, allowing models to learn effective decision boundaries directly from the coarsened representation.
    As a result, substantial performance can be retained without sophisticated methods, although such methods still achieve strong results.
    Only for Credit, the smallest and overall most challenging dataset, LLM-based imputation retains an advantage, indicating that pre-trained knowledge can compensate for limited training data even when distributions align.
    More generally, LLM-based methods perform better on Credit and Income than on Diabetes and Employment, suggesting stronger pre-training coverage for the former domains.

    \textbf{Benefits of Data Preparation:}
        The differences between Scenarios AnTr and AnTe compared to AnBo can be quantified by comparing the best-performing methods from \cref{fig:heatmap} against directly using anonymized data without preparation.
        While AnTr and AnTe achieve average F1 improvements of +89\% and +93\%, respectively, AnBo benefits only marginally and only for the Credit dataset.
        This further highlights the strong dependency on the deployment context, with AnBo clearly favoring no preparation.

\begin{figure}[t]
    \centering
    \includegraphics[width=0.9\linewidth]{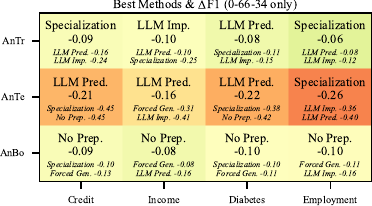}
    \caption{
        Heatmap of average F1 utility loss against original baseline for the three best preparation methods for the \texttt{0-66-34} distribution only (no original values).
    }
    \label{fig:heatmap-0-66-34}
\end{figure}

    \textbf{Absence of Original Data:}
        The privacy distribution \texttt{0-66-34}, in which no original values remain, represents the strongest privacy setting and tests the limits of all preparation methods.
        \Cref{fig:heatmap-0-66-34} shows the best-performing methods for each scenario-dataset combination under this extreme condition.
        Since most other approaches rely on observable original values, this setting naturally favors LLM-based methods that can leverage pre-trained knowledge.
        
        The general scenario dependence persists under \texttt{0-66-34} with AnTr and AnTe benefiting substantially from preparation, whereas AnBo again favors no preparation, now across all datasets.
        AnTe remains the most challenging scenario, with large F1 drops relative to the baseline that render it largely infeasible without original data.
        Although performance in AnTr and AnBo also declines compared to milder privacy settings, it remains feasible given the privacy level.
        In this extreme case, method selection shifts toward LLM-based prediction, which performs best overall.
        Even LLM-based imputation seems limited by the absence of original values, while specialization remains viable in some settings.
        This is likely due to the similarity of specialization to no preparation in the absence of original values, as categorical generalizations remain unchanged, while numerical features still benefit from interval midpoints.
        Overall, even without original data, data preparation methods can regain substantial utility (+89\% in AnTr, +120\% in AnTe), although the remaining gap to the baseline is particularly severe for AnTe.
        In contrast, AnBo continues to achieve strong results without preparation.
    
    \textbf{Forced Generalization:}
        The forced generalization method exhibits behavior that differs from its intended design.
        Although it aims to resolve train-test mismatches by enforcing a uniform generalization level, it consistently underperforms across scenarios due to discarding details.
        This is notable given that forced generalization often requires additional domain knowledge to define hierarchies, whereas other methods rely solely on observed anonymized values.
        Even in AnBo, where its behavior is comparable to no preparation, it offers no clear advantage over leaving the data unchanged.

\subsection{Aggregated Performance Results}\label{subsec:overall_results}

\begin{table*}[t]
\centering
\small
\setlength{\tabcolsep}{2.9pt}
\caption{Methods ranked by aggregation over scenarios, datasets, and distributions. On the left excluding and on the right including \texttt{0-66-34}, where no orignal data is left. We show F1 per scenario and dataset, and the mean F1 ($\overline{\text{\textbf{F1}}}$), precision ($\overline{\text{\textbf{P}}}$), and recall ($\overline{\text{\textbf{R}}}$). Best gives the percentage of aggregated cases where the method had the best F1. Note: Macro $\overline{\text{\textbf{F1}}}$ stems from per-class F1 scores not the macro $\overline{\text{\textbf{P}}}$ and $\overline{\text{\textbf{R}}}$. Abbr.: AnTr (Tr), AnTe (Te), AnBo (Bo), Credit (C), Income (I), Diabetes (D), Employment (E).}
\label{tab:overall_ranking}
\begin{tabular}{@{}l@{\hspace{4pt}}l ccc c ccc cccc c ccc c ccc cccc@{}}
\toprule
&
& \multicolumn{11}{c}{\textbf{Excluding \texttt{0-66-34}}}
& \multicolumn{11}{c}{\textbf{Including \texttt{0-66-34}}} \\
\cmidrule(lr){3-13}
\cmidrule(lr){14-24}

& 
& \multicolumn{4}{c}{\textbf{Aggregation}}
& \multicolumn{3}{c}{\textbf{Scenarios}}
& \multicolumn{4}{c}{\textbf{Datasets}}

& \multicolumn{4}{c}{\textbf{Aggregation}}
& \multicolumn{3}{c}{\textbf{Scenarios}}
& \multicolumn{4}{c}{\textbf{Datasets}} \\
\cmidrule(lr){3-6}
\cmidrule(lr){7-9}
\cmidrule(lr){10-13}
\cmidrule(lr){14-17}
\cmidrule(lr){18-20}
\cmidrule(lr){21-24}

& \textbf{Method}
& $\overline{\textbf{F1}}$
& $\overline{\textbf{P}}$
& $\overline{\textbf{R}}$
& \textbf{Best}
& \textbf{Tr}
& \textbf{Te}
& \textbf{Bo}
& \textbf{C}
& \textbf{I}
& \textbf{D}
& \textbf{E}
& $\overline{\textbf{F1}}$
& $\overline{\textbf{P}}$
& $\overline{\textbf{R}}$
& \textbf{Best}
& \textbf{Tr}
& \textbf{Te}
& \textbf{Bo}
& \textbf{C}
& \textbf{I}
& \textbf{D}
& \textbf{E} \\
\midrule

& Original Base.
& 0.77 & 0.78 & 0.77 & --
& 0.77 & 0.77 & 0.77
& 0.68 & 0.81 & 0.75 & 0.82
& 0.77 & 0.78 & 0.77 & --
& 0.77 & 0.77 & 0.77
& 0.68 & 0.81 & 0.75 & 0.82 \\
\midrule

\#1 & Specialization
& \textbf{0.68} & \textbf{0.70} & \textbf{0.69} & \textbf{42\%}
& \textbf{0.74} & \textbf{0.62} & \textbf{0.69}
& \textbf{0.59} & 0.70 & \textbf{0.69} & \textbf{0.76}
& \textbf{0.65} & \textbf{0.69} & \textbf{0.67} & \textbf{38\%}
& \textbf{0.72} & 0.56 & 0.67
& \textbf{0.57} & 0.65 & \textbf{0.66} & \textbf{0.74} \\

\#2 & LLM Imp.
& 0.65 & \textbf{0.70} & 0.66 & 17\%
& 0.73 & 0.56 & 0.65
& 0.58 & \textbf{0.74} & 0.63 & 0.64
& 0.62 & 0.67 & 0.65 & 15\%
& 0.71 & 0.52 & 0.64
& 0.54 & \textbf{0.71} & 0.60 & 0.64 \\

\#3 & LLM Pred.
& 0.61 & 0.63 & 0.63 & 4\%
& 0.66 & 0.59 & 0.59
& 0.49 & 0.69 & 0.61 & 0.66
& 0.61 & 0.62 & 0.63 & 10\%
& 0.66 & \textbf{0.58} & 0.58
& 0.49 & 0.69 & 0.61 & 0.64 \\

\#4 & Standard Imp.
& 0.61 & 0.66 & 0.64 & 10\%
& 0.73 & 0.47 & 0.62
& 0.56 & 0.71 & 0.58 & 0.57
& 0.55 & 0.59 & 0.61 & 8\%
& 0.66 & 0.43 & 0.57
& 0.52 & 0.64 & 0.53 & 0.53 \\

\#5 & No Prep.
& 0.49 & 0.53 & 0.57 & 27\%
& 0.42 & 0.35 & \textbf{0.69}
& 0.44 & 0.58 & 0.46 & 0.46
& 0.48 & 0.51 & 0.57 & 28\%
& 0.41 & 0.33 & \textbf{0.69}
& 0.43 & 0.56 & 0.46 & 0.46 \\

\#6 & Forced Gen.
& 0.43 & 0.45 & 0.54 & 0\%
& 0.34 & 0.32 & 0.64
& 0.41 & 0.47 & 0.48 & 0.37
& 0.44 & 0.44 & 0.55 & 0\%
& 0.34 & 0.32 & 0.64
& 0.34 & 0.53 & 0.46 & 0.42 \\

\bottomrule
\end{tabular}
\end{table*}
    
    Aggregating results across all scenarios, datasets, and privacy configurations yields the overall method ranking shown in \cref{tab:overall_ranking}.
    Our specialization method emerges as the strongest overall approach, which is even more evident in these aggregations than in \cref{fig:heatmap,fig:heatmap-0-66-34}.
    Although it primarily outperforms in scenarios AnTr and AnTe, it remains competitive against no preparation in AnBo.
    Across all settings, specialization achieves the lowest average F1 loss relative to the baseline, with a mean reduction of 0.09 when excluding and 0.12 when including the \texttt{0-66-34} distribution.
    It also exhibits the highest win rates, achieving the best F1 score in 42\% and 38\% of all aggregated combinations.
    As a result, specialization stands out as the most versatile method, delivering strong and consistent performance across scenarios, datasets, and privacy distributions.
    Even when including the unfavorable \texttt{0-66-34} setting, specialization does not fall behind competing approaches, despite being at a disadvantage compared to LLM-based methods that can leverage general pre-training knowledge in the absence of original values.
    LLM-based imputation ranks second overall, highlighting the benefit of external knowledge over standard imputation techniques.
    In contrast, while no preparation shines in the AnBo scenario, it performs poorly in AnTr and AnTe, where data preparation substantially improves utility.
    Consequently, no preparation and forced generalization occupy the lowest overall ranks when aggregating across all scenarios.

    \textbf{Precision and Recall:}
        In addition to F1 score, \cref{tab:overall_ranking} reports precision (P) and recall (R).
        Across most methods, the balance between precision and recall remains fairly stable.
        The notable exception is forced generalization, which exhibits disproportionately high recall at the expense of precision.
        A similar pattern emerges for the \texttt{0-66-34} distribution, where precision degrades considerably more than recall, while other distributions remain stable.
        This imbalance can be problematic, as correct predictions (high recall) are increasingly obscured by false positives (low precision).

    \textbf{Computational Trade-offs:} 
        Several methods introduce notable computational overhead, which is negligible in AnBo when no preparation is applied but becomes relevant in the other scenarios.
        Specialization increases the effective dataset size from $n$ to up to $O(n \cdot \tilde{v})$, depending on the number of retained variants $\tilde{v}$ and the prevalence of categorical anonymization. This leads to up to $\tilde{v}$-times the ML training or inference workload.
        LLM-based methods, in contrast, require a large number of LLM requests for anonymized records, even when batching is applied, resulting in increased runtime and potentially monetary costs for paid API tokens.
        Furthermore, additional costs may arise from the necessary LLM pre-training (or fine-tuning).
        Overall, LLM-based approaches incur higher overhead than specialization, with imputation taking roughly 40$\times$ and prediction about 8$\times$ longer on average under \texttt{66-17-17}.
        The additional computational cost of specialization can be justified by the utility gains in the AnTr and AnTe scenarios.

\begin{figure}[t]
    \centering
    \includegraphics[width=\linewidth]{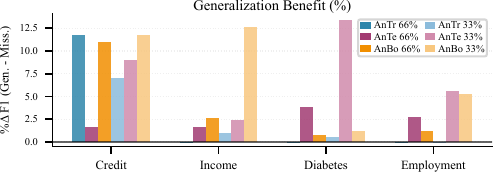}
    \caption{%
        Percentage change in best F1 score result when including generalized values (\texttt{66-17-17} and \texttt{33-33-34}) 
        compared to only setting values to missing (\texttt{66-0-34} and \texttt{33-0-67}). 
    }
    \label{fig:benefit}
\end{figure}

\subsection{Generalization Benefit}
    
    A key evaluation in our data model is whether allowing generalized values actually increases utility over simply using missing values.
    In \cref{fig:benefit}, we compare the best result from distributions that are a mix of generalized and missing values against a contrastive setting with the respective amount missing.
    Including generalizations, which retain partial information about the underlying originals, consistently yields better or similar results than complete suppression.
    Benefits are most evident for our specialization method, which explicitly exploits the hierarchical structure, as well as for LLM-based imputation, which can use generalized values as contextual information.
    Although the magnitude of the generalization benefit differs across scenarios and datasets, it never exhibits a perceptible negative impact. Instead, each dataset sees some utility gains.
    
    The Credit dataset is most responsive across all cases, while in the AnTr scenario, generalization appears less beneficial for the remaining datasets, with utility largely unchanged.
    In contrast, for Ante and AnBo, all datasets consistently benefit from including generalized values, particularly in the lower original setting (33\%).
    
    Overall, these results demonstrate that generalization is less harmful than suppression and can actively regain utility for downstream tasks.
    The effect seems stronger for smaller datasets such as Credit, where every data point matters.
    However, for these benefits to materialize, data preparation methods must be capable of leveraging the additional information provided by generalized values.

\subsection{Privacy Results}

\begin{table}[t]
\centering
\caption{Privacy metrics by setting and dataset as marketer risk ($R_m$$\downarrow$), $k$-anonymity ($k$$\uparrow$), and their aggregated mean.}
\label{tab:privacy_dis_data}
\small
\setlength{\tabcolsep}{3.3pt}
\begin{tabular}{@{}lcccccccc|cc@{}}
\toprule
\textbf{Distrib.} & \multicolumn{2}{c}{\textbf{Credit}} & \multicolumn{2}{c}{\textbf{Income}} & \multicolumn{2}{c}{\textbf{Diabetes}} & \multicolumn{2}{c}{\textbf{Employ.}} & \multicolumn{2}{c}{\textbf{mean}} \\
\cmidrule(lr){2-3} \cmidrule(lr){4-5} \cmidrule(lr){6-7} \cmidrule(lr){8-9} \cmidrule(lr){10-11}
& $R_m$ & $k$ & $R_m$ & $k$ & $R_m$ & $k$ & $R_m$ & $k$ & $R_m$ & $k$ \\
\midrule
\texttt{1-0-0}     & 1.00 & 1.0   & 1.00 & 1.0   & 0.98 & 1.0   & 0.51 & 1.0   & 0.87 & 1.0 \\
\texttt{66-17-17}  & 0.96 & 1.0   & 0.98 & 1.0   & 0.95 & 1.0   & 0.58 & 80.4  & 0.87 & 20.9 \\
\texttt{66-0-34}   & 0.96 & 1.0   & 0.98 & 1.0   & 0.94 & 1.0   & 0.55 & 80.1  & 0.86 & 20.8 \\
\texttt{33-33-34}  & 0.96 & 1.0   & 0.90 & 1.0   & 0.81 & 1.0   & 0.58 & 38.2  & 0.81 & 10.3 \\
\texttt{33-0-67}   & 0.95 & 1.0   & 0.79 & 1.0   & 0.74 & 1.0   & 0.54 & 35.4  & 0.76 & 9.6 \\
\texttt{0-66-34}   & 0.73 & 119.3 & 0.41 & 5.4K  & 0.26 & 8.4K  & 0.12 & 46K   & 0.38 & 15K \\
\midrule
\textbf{mean}      & 0.93 & 20.7  & 0.84 & 0.9K  & 0.78 & 1.4K  & 0.48 & 7.7K  & 0.76 & 2.5K \\
\bottomrule
\end{tabular}
\end{table}

    As given in \cref{sec:experiments}, we report marketer re-identification risk ($R_m$) and achieved $k$-anonymity in \cref{tab:privacy_dis_data}. Across all datasets, privacy outcomes are primarily driven by the chosen privacy distribution rather than the applied data preparation method, a pattern reflected in the method-wise aggregation shown in \cref{tab:privacy_methods}.

    \setlength{\columnsep}{10pt}
    \begin{wraptable}{r}{3.3cm}
    \vspace{-0.8\intextsep}
    \centering
    \caption{Privacy by method (excl. \texttt{0-66-34}).}
    \label{tab:privacy_methods}
    \small
    \setlength{\tabcolsep}{3pt}
    \begin{tabular}{@{}lcc@{}}
    \toprule
    \textbf{Method} & $\boldsymbol{R_m}$ & $\boldsymbol{k}$ \\
    \midrule
    Original Base. & 0.87 & 1.0 \\
    \midrule
    No Prep. & 0.92 & 1.0 \\
    Standard Imp. & 0.73 & 33.1 \\
    Forced Gen. & 0.81 & 1.0 \\
    Specialization & 0.91 & 1.0 \\
    LLM Imp. & 0.79 & 31.7 \\
    LLM Pred. & 0.92 & 1.0 \\
    \bottomrule
    \end{tabular}
    \vspace{-\intextsep}
    \vspace{3pt}
    \end{wraptable}

    For all configurations that retain original values, $R_m$ remains relatively high, with a noticeable reduction only once the fraction of original values drops to 33\%. Less anonymized settings (\texttt{66-17-17} and \texttt{66-0-34}) exhibit mean $R_m$ values comparable to the original baseline (0.87). Correspondingly, the achieved $k$-anonymity is 1 for Income, Diabetes, and Credit, indicating that many records remain unique under our conservative assumption that all attributes are treated as quasi-identifiers.
    Employment constitutes a partial exception with larger $k$-values even at moderate privacy distributions of 66\% originals, showing that dataset-specific characteristics also play a key role.
    A qualitatively different behavior is observed for the extreme \texttt{0-66-34} configuration, in which no original values remain. In this case, $R_m$ drops sharply across all datasets (mean $R_m = 0.38$), while $k$-anonymity increases by orders of magnitude, reaching up to $k = 46\text{K}$ for Employment. This highlights the strong aggregation effect induced by the complete removal of original attribute values, which effectively eliminates record-level uniqueness.

    \Cref{tab:privacy_methods} summarizes privacy metrics for our preparation methods, excluding \texttt{0-66-34} as it would dominate the aggregations, and shows that differences between methods are moderate. Using anonymized data itself without any data preparation surprisingly yields the worst results ($R_m = 0.92$, $k = 1.0$), with identical values for LLM prediction that operates on this data. This suggests that classical privacy metrics may not adequately capture the privacy properties of heterogeneous, user-driven anonymization, as generalized values and missingness can introduce record-level uniqueness when treated as exact identifiers.
    This notion is preserved by specialization, despite reversing generalization, since expanding generalized values into multiple variants creates distinct record profiles where each variant differs slightly in its reconstructed values and thereby increases uniqueness under syntactic re-identification metrics.
    In contrast, standard and LLM-based imputation yield slightly better privacy, likely due to imputing more homogeneous attribute values and thereby increasing equivalence class sizes.
    
    Overall, none of the evaluated data preparation methods oppose or reverse user-driven anonymization and instead maintain comparable privacy levels. 
    Instead, the observed differences in privacy metrics are primarily determined by the user-driven privacy distribution of an anonymized dataset and the remaining fraction of original values.
    Importantly, our methods that improve utility (e.g., specialization) do not compromise the measured privacy of the anonymized data under $R_m$ and $k$-anonymity.

\section{Discussion} 
\label{sec:discussion}

We discuss our findings on data preparation for tabular datasets with heterogeneous privacy transformations.
All methods operate directly on anonymized data by potentially altering its representation to regain downstream utility without modifying the ML model.

\textbf{Utility:}
    The effectiveness of data preparation is highly scenario dependent.
    The AnTr scenario is the most forgiving setting, with only minor F1-score degradation, especially under our proposed specialization method.
    AnTe also benefits substantially from data preparation but is the most challenging scenario. In particular, when no original values remain, learning becomes nearly infeasible.
    
    In contrast, AnBo requires no data preparation to retain strong utility.
    While preparation still works in this setting, not preparing yields the same results and avoids computational overhead.
    Here, consistent anonymization across training and inference allows models to learn directly from coarsened representations, while reconstruction-based methods may introduce unnecessary noise.
    Further, anonymization generally does not introduce errors, with generalized values still conveying correct, albeit coarsened, information.
    The AnBo setting thereby defies the assumption that heterogeneous values must be inherently problematic, as no preparation that preserves heterogeneity performs well, whereas forced generalization, which explicitly enforces homogeneity, performs poorly.
    
    Direct LLM prediction outperforms in AnTe when no original data is available, leveraging pre-training to compensate for unobservable test feature domains.
    Specialization remains competitive but is inherently limited without original values, similar to other imputation-based approaches.
    Against baseline imputation, specialization yields a 12\% to 18\% relative improvement in mean F1.
    
    Overall, specialization is the most robust and versatile preparation strategy, performing consistently well across scenarios and with minimal assumptions.
    It is effective for both generalized and missing values, while it clearly benefits from retaining generalized values rather than full suppression.
    Finally, if the deployment scenario is known, we recommend no preparation for AnBo and specialization for AnTr and AnTe.
    However, if the scenario is unknown or may change, specialization is the safest default choice.

\textbf{Privacy:}
    A concern for our data preparation on anonymized data is whether it compromises privacy by reversing anonymization.
    Our results show that our methods do not meaningfully increase re-identification risk.
    Instead, privacy is largely driven by the data distribution and particularly the share of original values retained.
    Notably, unprepared anonymized data appears less private than the original baseline under marketer risk and $k$-anonymity, even though it is conceptually more privacy-preserving.
    These metrics treat generalized values as exact identifiers and fail to capture set-valued semantics.
    As a result, increased heterogeneity can inflate record uniqueness without actually weakening the intended privacy protection, especially under our conservative assumption that all attributes are quasi-identifiers.
    While classical metrics remain useful, they should be interpreted cautiously for heterogeneous, user-driven anonymization.
    It is important to note that such privacy transformations do not provide formal guarantees and future work would therefore benefit from introducing metrics better aligned with informing user-driven privacy decisions.

\textbf{Scalability:}
    We compare the computational overhead of LLM and specialization in \cref{subsec:overall_results}, showing significantly higher runtimes for LLM-based methods.
    Our experiments represent a near worst-case setting for specialization, as all features are treated as sensitive and anonymized column-wise, leading to mixed records and, therefore, numerous variants per record.
    Still, our filtering to fewer candidates (e.g., $\tilde{v}=2$) keeps dataset growth manageable, with a maximum increase of $\tilde{v}$-times the original ML compute, in contrast to other dataset-level or generative augmentation methods, which are known to incur substantially higher costs~\cite{mumuni2022data,wang2025comprehensive}.
    However, for datasets with many categorical features and very large value domains, even the filtering step can introduce non-negligible overhead.
    In practice, usually only a subset of attributes is treated as sensitive, which substantially improves scalability~\cite{khan2025shades}.
    Reductions in feature space are also possible by prioritizing the specialization of features based on importance scores or by consolidating large categorical domains during data collection.
    Exploring such pre-filtering strategies is a promising direction for future work.

\textbf{Assumptions:}
    We adopt conservative assumptions regarding domain knowledge without access to full feature hierarchies or external knowledge.
    Instead, we rely solely on values observed after anonymization, without ground truth, auxiliary data, or pre-training.
    Our specialization only requires linking generalized values to observed originals, while missing values are directly observable.
    In practice, domain hierarchies are often known or inferable, and future work could leverage LLMs to approximate them, especially when no original values remain.
    We further assume column-wise rather than user-specific privacy distributions, as this would introduce additional assumptions about the user population and may have a disparate influence due to the relative importance of individual records.
    We apply these privacy transformations to all attributes, removing the possible bias from selecting which attributes might be sensitive for a user.
    Our evaluation relies on synthetically generated datasets, which is a common limitation due to the lack of real-world data with privacy ground truth.
    Finally, rather than exact reconstruction of the original data, our goal is to improve downstream utility under privacy constraints through data preparation.

\textbf{Transferability:}
    Our findings extend to tasks beyond classification that involve statistical analysis of anonymized data.
    In addition to downstream machine learning, this includes problems that operate on aggregated or approximate outputs, such as aggregation, filtering, and grouping queries.
    This transfer follows from the fact that our approach targets feature representations at the data level rather than the downstream application itself.
    Since heterogeneous feature domains also arise from data integration, schema evolution, or inconsistent data collection, our methods remain applicable beyond privacy-driven settings.
    Furthermore, as demonstrated, they directly handle incomplete datasets caused solely by missing data.

\textbf{Relation to Prior Work:}
    Our findings complement prior work on data quality and ML performance from \cref{sec:background,tab:datasets} by extending to mixed privacy transformations and proposing appropriate preparation strategies.
    Comparable to earlier studies~\cite{mohammed2025dq,khan2025shades}, we observe substantial degradation from anonymization under mismatched training and inference conditions. However, we demonstrate that appropriate preparation can regain much of the lost utility.
    Compared to settings with a smaller sensitive attribute set (sex, age, and race)~\cite{khan2025shades}, our conservative privacy assumptions represent a stricter case. Relaxing our set from all attributes to only a subset is likely to improve utility, scalability, and privacy metrics.
    Unlike data cleaning systems that assume some access to mostly clean data~\cite{reis2024lopster} for training, we operate exclusively on anonymized inputs and treat generalized values as valid constraints rather than errors.
    Preserving and exploiting these constraints enables utility gains that standard imputation methods often cannot achieve.
\section{Conclusion}
\label{sec:conclusion}

Data collection under user-driven privacy leads to tabular datasets in which original, generalized, and missing values may co-exist within attributes and records.
This poses a challenge for standard ML pipelines that consequently suffer from inconsistent feature domains and introduced distribution shifts.
In this work, we propose and evaluate data preparation methods to support learning under such heterogeneous data representations across realistic deployment scenarios.
Our results show that generalized values consistently preserve more utility than pure suppression, that preparation effectiveness is strongly scenario dependent, and that mismatches between handling training and inference representations are a primary source of performance loss.
We further find that consistent anonymization across training and inference can eliminate the need for preparation altogether.
Across all settings, our proposed granularity-based imputation using specialization emerges as the most robust preparation strategy, regaining substantial classification utility under limited assumptions and without compromising measured privacy.
These findings demonstrate that effective learning under user-driven privacy does not require modifying ML models or reversing anonymization.
Instead, significant utility can be recovered by preparing anonymized data accordingly, making appropriate data preparation key for practical, privacy-aware ML.

Future work may address gathering real-world datasets or extend to other downstream ML and analytical tasks on comparable data representations. They would also benefit from developing more suitable and informative privacy metrics for user-driven privacy.

\begin{acks}
    The authors acknowledge the financial support by the Federal Ministry of Education and Research of Germany and by the Sächsische Staatsministerium für Wissenschaft, Kultur und Tourismus for ScaDS.AI.
    Computations for this work were done (in part) using resources of the Leipzig University Computing Centre.
    This work was partially funded by Universities Australia and the German Academic Exchange Service (DAAD) grant 57701258.
\end{acks}

\onecolumn \begin{multicols}{2} 

\bibliographystyle{ACM-Reference-Format}
\bibliography{paper.bib, sample.bib} 

\end{multicols}

\balance  

\end{document}